\documentclass{article}
\usepackage{spconf,amsmath,graphicx}

\usepackage{enumitem}
\setlist{nosep, leftmargin=14pt}

\usepackage{mwe} 

\usepackage{multirow,commath}
\usepackage{subfig}
\usepackage{hhline}
\usepackage{xkeyval,xcolor}
\makeatletter
\newlength{\sfp@hseplen}\newlength{\sfp@vseplen}
\define@cmdkey{subfigpos}[sfp@]{pos}[ul]{}
\define@cmdkey{subfigpos}[sfp@]{font}[\small]{}
\define@cmdkey{subfigpos}[sfp@]{vsep}[5pt]{\setlength{\sfp@vseplen}{\sfp@vsep}}
\define@cmdkey{subfigpos}[sfp@]{hsep}[5pt]{\setlength{\sfp@hseplen}{\sfp@hsep}}
\newcommand{\subfigimg}[3][,]{%
  \setkeys{Gin,subfigpos}{pos,font,vsep,hsep,#1}
  \setbox1=\hbox{\includegraphics{#3}}
  \ifnum\pdfstrcmp{\sfp@pos}{ul}=0
    \leavevmode\rlap{\usebox1}
    \rlap{\hspace*{\sfp@hsep}\raisebox{\dimexpr\ht1-\sfp@vsep}{\sfp@font{#2}}}
    \phantom{\usebox1}
  \else\ifnum\pdfstrcmp{\sfp@pos}{ur}=0
    \leavevmode\usebox1
    \llap{\raisebox{\dimexpr\ht1-\sfp@vsep}{\sfp@font{#2}}\hspace*{\sfp@hsep}}
  \else\ifnum\pdfstrcmp{\sfp@pos}{lr}=0
    \leavevmode\usebox1
    \llap{\raisebox{\sfp@vsep}{\sfp@font{#2}}\hspace*{\sfp@hsep}}
  \else
    \leavevmode\rlap{\usebox1}
    \rlap{\hspace*{\sfp@hseplen}\raisebox{\sfp@vsep}{\sfp@font{#2}}}
    \phantom{\usebox1}
  \fi\fi\fi
}
\makeatother

\begin{document}
\twocolumn[  
\begin{@twocolumnfalse}
	This article has been accepted for publication. Citation information: DOI 10.1109/ISBI52829.2022.9761423, 2022 IEEE 19th International Symposium on Biomedical Imaging (ISBI). See https://ieeexplore.ieee.org/abstract/document/9761423 for the published article.
	
	
\end{@twocolumnfalse}
]

\title{Graph-based Small Bowel Path Tracking with Cylindrical Constraints}
%
\name{Seung Yeon Shin \qquad Sungwon Lee \qquad Ronald M. Summers}
\address{Imaging Biomarkers and Computer-Aided Diagnosis Laboratory\\Radiology and Imaging Sciences, National Institutes of Health Clinical Center, USA}
%
%
%
%
%
%
\maketitle
\begin{abstract}
We present a new graph-based method for small bowel path tracking based on cylindrical constraints. A distinctive characteristic of the small bowel compared to other organs is the contact between parts of itself along its course, which makes the path tracking difficult together with the indistinct appearance of the wall. It causes the tracked path to easily cross over the walls when relying on low-level features like the wall detection. To circumvent this, a series of cylinders that are fitted along the course of the small bowel are used to guide the tracking to more reliable directions. It is implemented as soft constraints using a new cost function. The proposed method is evaluated against ground-truth paths that are all connected from start to end of the small bowel for 10 abdominal CT scans. The proposed method showed clear improvements compared to the baseline method in tracking the path without making an error. Improvements of $6.6\%$ and $17.0\%$, in terms of the tracked length, were observed for two different settings related to the small bowel segmentation.
\end{abstract}
\section{Introduction}\label{sec:intro}

The small bowel is a major digestive organ. It is about 6 meters long and extends from the pylorus to the ileocecal junction~\cite{gs20}.
Being highly convoluted, it has abundant contact between parts of itself as well as with surrounding organs.
Its appearance also differs dramatically along its length since bowel contents are locally highly variable.

Computed tomography (CT) is frequently used for small bowel inspection since it is fast and patient friendly~\cite{murphy14}. Interpretation of 3D CT scans is done by radiologists, and is laborious and time-consuming.

In an effort to assist this, automatic methods that identify the small bowel have been developed~\cite{zhang13,oda20,shin20,shin21_sb_segm,oda21,harten21,shin21_sb_track}. The previous works on small bowel segmentation have improved their clinical applicability in different ways, which are by training a neural network with sparse annotation~\cite{oda20}, by incorporating a shape prior~\cite{shin20}, or by introducing a new domain adaptation technique~\cite{shin21_sb_segm}. Although the segmentation provides essential information for downstream tasks such as detecting abnormalities in the small bowel by distinguishing it from the surroundings, it is deficient in understanding the whole structure due to the aforementioned contact issue.

Recent research on small bowel path tracking has been performed to complement the shortcomings of the segmentation~\cite{oda21,harten21,shin21_sb_track}. A unique neural network is trained to predict the distance from the small bowel centerlines for each voxel~\cite{oda21} or to decide a direction in which the tracker moves~\cite{harten21}. Despite the promising results presented, they each have limitations. In \cite{oda21}, air-inflated bowels, where the walls are more distinguishable from the lumen than with other internal material, were used for qualitative evaluation. In \cite{harten21}, the tracking was performed for each segment of the small bowel under a restricted field-of-view of cine-MR scans. Therefore, both of the methods are not guaranteed to work for tracking the entire course of the small bowel from routine CT scans. It was attempted in \cite{shin21_sb_track}, where the minimum cost path between given start and end nodes is computed on a bowel graph that is constructed based on the wall detection. Achieving a high recall throughout the entire small bowel with the help of the must-pass nodes, it is vulnerable to errors from the wall detection.

In this paper, we thus propose an enhanced graph-based method for small bowel path tracking. It is built upon the method of \cite{shin21_sb_track}, but with a new cost function that is facilitated by local cylinder fitting.
The proposed framework will be explained with an emphasis on the cylinder fitting in the following section.

\section{Method}\label{sec:method}

\begin{figure*}[t]
    \centering
    \includegraphics[width = 0.95\textwidth]{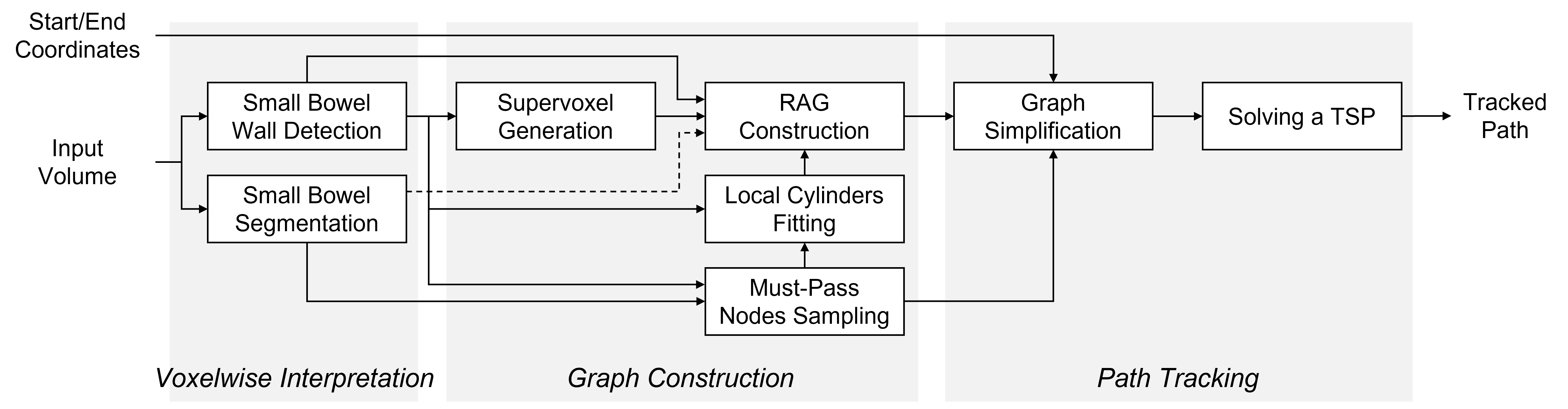}
    \caption{Pipeline of the proposed method. A region adjacency graph (RAG) is constructed based on small bowel wall detection and segmentation. A set of sampled must-pass nodes are used to better cover the entire course of the small bowel. Also, locally fitted cylinders are used to encourage the tracking to follow more reliable directions. Finally, the optimal path is achieved by solving a travelling salesman problem (TSP) on a simplified graph, composed of the start, end, and the must-pass nodes.}
    \label{fig:pipeline} 
\end{figure*} 

\subsection{Dataset}\label{ssec:dataset}

Our dataset is composed of 30 intravenous and oral contrast-enhanced abdominal CT scans. A positive oral contrast agent, Gastrografin, was used. The scans were done during the portal venous phase.
All scans were resampled to have isotropic voxels of $2^3mm^3$.
Then, they were manually cropped along the z-axis to include from the diaphragm through the pelvis.

The dataset includes two types of ground-truth (GT) labels, namely, segmentation and path of the small bowel.
A radiologist with 12 years of experience used ``Segment Editor" module in 3DSlicer~\cite{fedorov12} to acquire the GTs, which cover the entire small bowel including the duodenum, jejunum, and ileum. While all 30 scans have a GT segmentation, GT path is achieved for a subset of 10 scans because of the exceptionally time-consuming nature of this manual task (one full day for each scan). An interpolated curve that connects a series of points manually placed inside the small bowel, from start to end, is drawn. We note that our GT path is all connected for the entire small bowel while a set of segments that are $88mm$ long on average was used in \cite{harten21}. This subset is used for the evaluation of the proposed path tracking method. The remaining 20 scans, which we will call the segmentation training set, are used to train a network for segmentation prediction.

\subsection{Voxelwise Interpretation of the Small Bowel}\label{ssec:vxl_interp}

The overall pipeline of the proposed method is presented in Fig.~\ref{fig:pipeline}.
In our dataset, the lumen appears brighter than the bowel walls due to the oral contrast. To prevent the tracking from penetrating the walls around the contact between different parts of the small bowel, we detect the walls by finding valleys in an input volume. The Meijering filter~\cite{meijering04} is used.

The small bowel segmentation is performed in parallel with the wall detection. It is not only to restrict the graph construction within the small bowel but also for must-pass nodes sampling. To this end, we use the 3D U-Net~\cite{cicek16} architecture used in \cite{shin20}. It is trained using the segmentation training set. We note that this is the only part where training is involved. No GT path is required to establish the proposed method while the previous methods~\cite{oda21,harten21} do.

\subsection{Graph Construction}

We generate supervoxels from the output of the wall detection by Adaptive-SLIC~\cite{achanta12}. Thus, the generated supervoxels are expected to well adhere to boundaries between the lumen and the walls. $S=\{s_i\}_{i=1}^{N}$ is the set of the supervoxels, where $N$ is the number of the supervoxels.

A region adjacency graph (RAG), $G=(V,E)$, is constructed by taking each supervoxel $s_i$ as a node $v_i$.
$V=\{v_i\}_{i=1}^{N}$ is the set of the nodes and $E$ is a set of edges. An edge $e_{i,j}$ is created between two adjacent supervoxels $s_i$, $s_j$.

\subsubsection{Must-Pass Nodes Sampling}

In \cite{shin21_sb_track}, it is observed that simply finding the shortest path between given start and end nodes on a graph produces a trivial solution with many short-cuts through the bowel wall due to the imperfection of the wall detection. A set of must-pass nodes were used to enable the search to better cover the entire course of the small bowel. We use the same scheme in this work. First, the Euclidean distance transformation is computed from the logical OR output of the inverted small bowel segmentation and the binarized wall detection. Thus, each voxel value indicates the Euclidean distance to the nearest wall. Finally, the must-pass nodes are sampled as local peaks on the distance map. Two requirements on the minimum value of peaks, $\theta_v^{peak}$, and on the minimum distance between peaks, $\theta_d^{peak}$, are fulfilled during the sampling. Thus, they are possibly the innermost voxels that are properly distributed within the lumen. The set of must-pass nodes is denoted by $V^{mp}$, where $V^{mp} \subset V$ and $|V^{mp}|\ll|V|$.

\subsubsection{Local Cylinders Fitting}

\begin{figure*}[t]
	\centering
	\begin{minipage}{1\textwidth}
        \subfigimg[width=0.2\textwidth, pos=ll, font=\color{white}]{A}{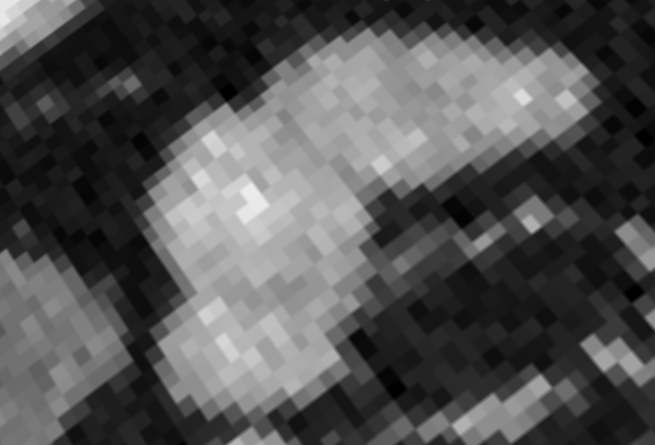}
        \hspace{-0.17cm}
        \subfigimg[width=0.2\textwidth, pos=ll, font=\color{white}]{B}{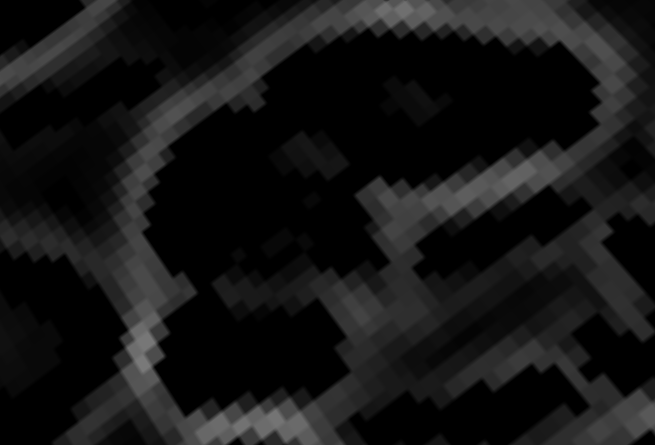}
        \hspace{-0.17cm}
        \subfigimg[width=0.2\textwidth, pos=ll, font=\color{white}]{C}{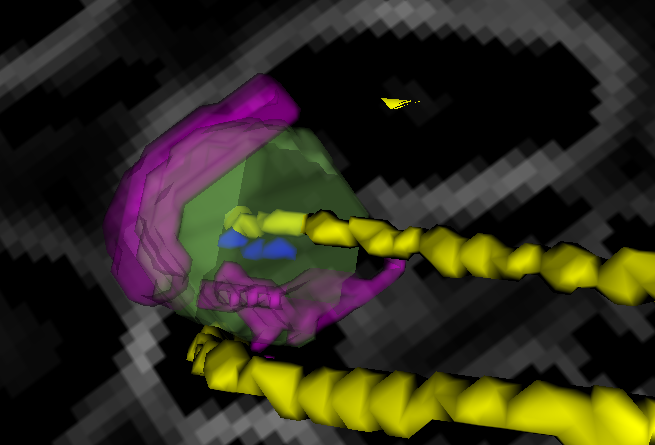}
        \hspace{-0.17cm}
        \subfigimg[width=0.2\textwidth, pos=ll, font=\color{black}]{D}{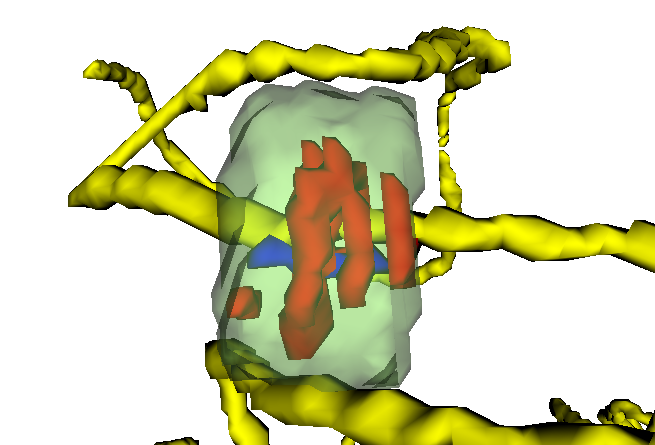}
        \hspace{-0.17cm}
        \subfigimg[width=0.2\textwidth, pos=ll, font=\color{black}]{E}{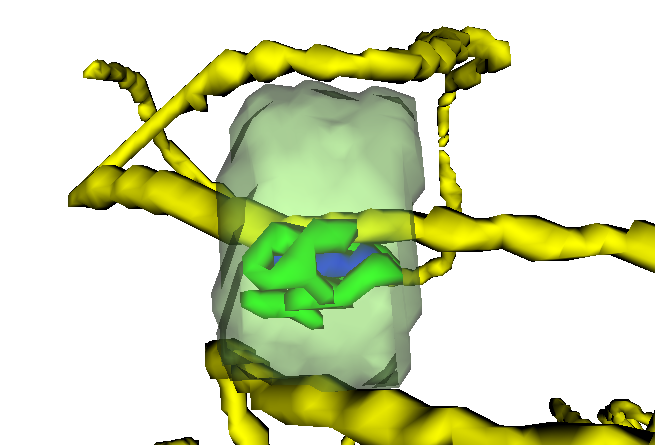}
    \end{minipage}
	\caption{Local cylinder fitting. Only one cylinder of interest is visualized. (A) A zoomed image slice of the input volume. (B) Corresponding wall detection map. (C) A cylinder (transparent green) estimated from the surrounding binarized wall detection (purple). The cylinder axis and ground-truth path are also shown in blue and yellow, respectively. (D)(E) The cylinder of (C) is shown from a different viewpoint to better explain its role. Graph edges that are located within the cylinder are colored according to their respective cost, $C_{i,j}^{cyl}$. Only edges that have a cost higher than 0.8 (red) or lower than 0.2 (green) are drawn.}
	\label{fig:cyl}
\end{figure*}

In \cite{shin21_sb_track}, the edge cost relies solely on values between supervoxels on the wall detection map. Therefore, it is susceptible to errors from the wall detection. While false positives within the lumen could cause a locally winding tracked path, false negatives could cause crossing over the walls.

To prevent the tracking from changing the direction too freely according to nearby detected walls that might include errors, we predict more reliable directions by looking wider in the wall detection. We regard the small bowel as piecewise cylindrical, and fit a cylinder for the location of each must-pass node by the RANSAC algorithm\footnote{https://leomariga.github.io/pyRANSAC-3D}~\cite{fischler81}, which is robust to outliers. 
Parameters defining an infinite height cylinder, center, axis, and radius, are estimated from a local patch of the binarized wall detection. A fixed height is then used for every cylinder (Fig.~\ref{fig:cyl} (C)).

The edge cost $C_{i,j}$ between two adjacent supervoxels $s_i$, $s_j$ is then defined as follows:
\begin{equation}
    \label{eq:edge_cost}
    C_{i,j} = C_{i,j}^{wall} + \lambda C_{i,j}^{cyl},
\end{equation}
where $C_{i,j}^{wall}$ is directly based on the values in the wall detection map, and $C_{i,j}^{cyl}$ is based on the fitted cylinders. $\lambda$ is the relative importance of $C_{i,j}^{cyl}$. $C_{i,j}^{wall}$ is defined as the average value, in the wall detection map, along the boundary between $s_i$ and $s_j$. Thus, it would be high for neighboring supervoxels that have a boundary of the lumen and the wall in between.

On the other hand, $C_{i,j}^{cyl}$ is defined as follows:

\begin{equation}
    \label{eq:edge_cost_cyl}
    C_{i,j}^{cyl} = 1 - \left |\frac{\overrightarrow{p(i)p(j)} \cdot \vec{a}}{\norm{\overrightarrow{p(i)p(j)}}\norm{\vec{a}}} \right |,
\end{equation}
where $\overrightarrow{p(i)p(j)}=p(j)-p(i)$ is the vector between the centroids of supervoxels $s_i$, $s_j$, and $\overrightarrow{a}$ is the axis vector of the local cylinder, within which $s_i$ and $s_j$ are located. $p(i)$ gives the centroid of supervoxel $s_i$. This cost favors edges that are parallel with the cylinder axis and disfavors perpendicular edges, as illustrated in Fig.~\ref{fig:cyl}. 0.5 is used as the default value of $C_{i,j}^{cyl}$ for edges that are not located within any local cylinder.

\subsection{Path Tracking}

A start node $v_{st}$ and an end node $v_{ed}$ are designated on the built graph $G$ by providing coordinates for each on an input CT volume.
Then, the optimal path that starts from $v_{st}$, and passes by the must-pass nodes $V^{mp}$, and ends at $v_{ed}$ is computed. The travelling salesman problem (TSP) based approach that is proposed in \cite{shin21_sb_track} is used to achieve the solution.

To this end, a simplified graph $G'=(V',E')$, composed of the start, end, and the must-pass nodes, is constructed, on which a TSP solution is computed. $V'=\{v'_m\}_{m=1}^{|V^{mp}|+2}=\{v_{st},v_{ed}\} \cup V^{mp}$ is a new set of nodes and $E'$ is a new set of edges. An edge is created for every pair of nodes in $G'$ and the edge cost $C_{m,n}'$ between nodes $v'_m$ and $v'_n$ is defined as:

\begin{equation}
    \label{eq:edge_cost_tsp}
    C_{m,n}'= 
    \begin{cases}
        \frac{cost(\Gamma^*_G(v'_m,v'_n))}{M}, & \text{if } d(v'_m,v'_n) \leq \delta\\
        \frac{d(v'_m,v'_n)}{\delta}, & \text{otherwise},
    \end{cases}
\end{equation}
where $\Gamma^*_G = \operatorname*{arg\,min}_{\Gamma_G} \sum_{e_{i,j} \in \Gamma_G} C_{i,j}$ is the shortest path found by the Dijkstra’s algorithm on $G$. The function $cost(\cdot)$ gives the total cost of the argument path. It is normalized by the maximum value, $M=\max_{\substack{(p,q) \in E' \\ d(v'_p,v'_q) \leq \delta}}  cost(\Gamma^*_G(v'_p,v'_q))$. $d(v'_m,v'_n)$ denotes the Euclidean distance between $v'_m$ and $v'_n$, and $\delta$ is the threshold on it. Depending on the Euclidean distance, the cost $C_{m,n}'$ is defined either by the shortest path cost or by the \emph{computationally lighter} Euclidean distance itself.  It is based on an assumption that a pair of nodes that are far from each other by the Euclidean distance would not be direct neighbors in a computed path.

The TSP solution that starts and ends at different nodes can be achieved by adding a dummy node that has zero-cost edges with the start and end nodes, and infinity-cost edges with the remaining nodes. The nearest fragment operator~\cite{ray07} is used to solve the TSP.

\begin{figure*}[t]
	\centering
	\begin{minipage}{1\textwidth}
        \subfigimg[width=0.2\textwidth, pos=ll, font=\color{black}]{A}{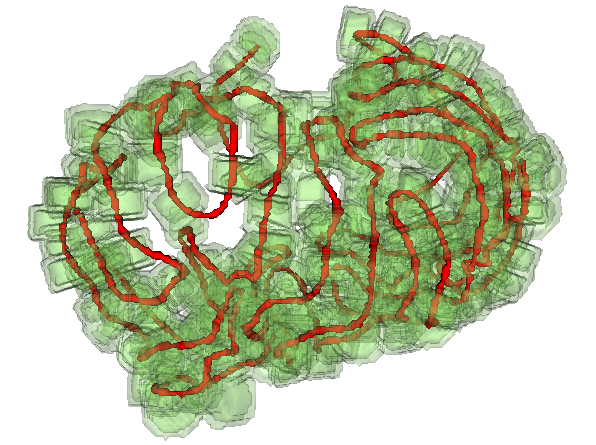}
        \hspace{-0.17cm}
        \subfigimg[width=0.2\textwidth, pos=ll, font=\color{black}]{B}{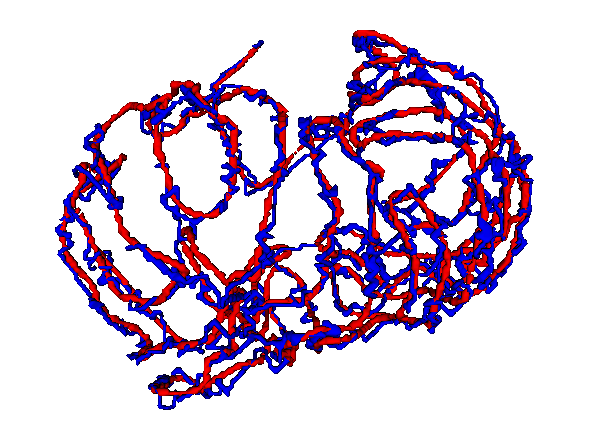}
        \hspace{-0.17cm}
        \subfigimg[width=0.2\textwidth, pos=ll, font=\color{black}]{C}{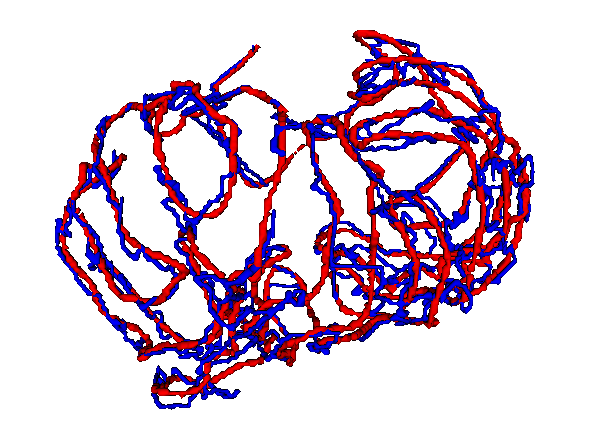}
        \hspace{-0.17cm}
        \subfigimg[width=0.2\textwidth, pos=ll, font=\color{white}]{D}{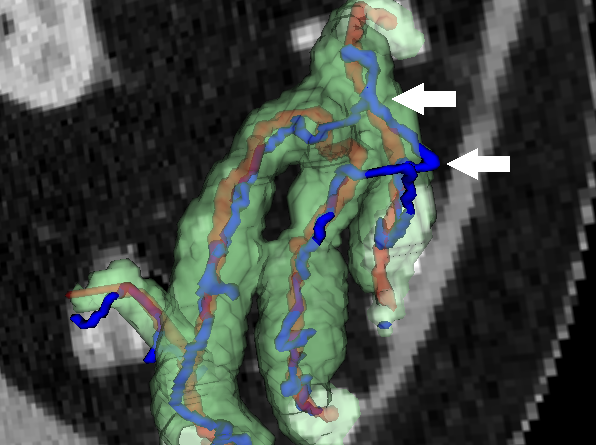}
        \hspace{-0.17cm}
        \subfigimg[width=0.2\textwidth, pos=ll, font=\color{white}]{E}{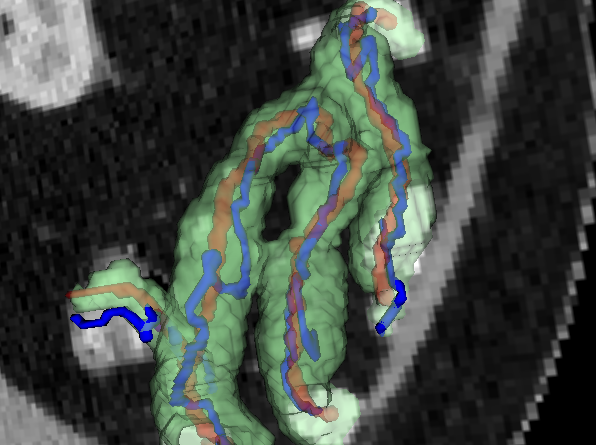}
    \end{minipage}
	\caption{Example path tracking results. (A) Fitted local cylinders (green) overlayed on the ground-truth (GT) path (red). (B) Result corresponding to `TSP' in Table~\ref{tab:quan_res}. (C) Result of the proposed method. In (B) and (C), each result (blue) is compared with the GT (red). (D)(E) A part of the tracked paths corresponding to (B) and (C), respectively. The areas indicated by arrows show the typical error of crossing the walls. The small bowel segmentation (green) is overlayed to highlight the contact issue, which can cause the aforementioned error.}
	\label{fig:qual_res}
\end{figure*}

\subsection{Evaluation Details}\label{ssec:eval_details}

The number of the supervoxels, $N$, was set variably according to the size of the input volume to make each supervoxel have a desired volume of $216mm^3$ in the Adaptive-SLIC~\cite{achanta12}. The compactness was set as $0.01$. We used $\theta_v^{peak}=3mm$ and $\theta_d^{peak}=6mm$ for must-pass nodes sampling.

Local patches of size $36^3mm^3$ were used for cylinder fitting. The cylinders were generated with a fixed height of $18mm$. 
The number of the RANSAC iterations and the distance threshold for inlier were set as $50,000$ and $1mm$, respectively. According to the statistics on the small bowel~\cite{cronin10}, the estimated cylinder radius was constrained to be within a range of $[7.04mm, 15.28mm]$. $\lambda=1$ was used for Eq.~(\ref{eq:edge_cost}), and $\delta=50mm$ was used for Eq.~(\ref{eq:edge_cost_tsp}). All the values except for the range of cylinder radius were chosen by experiment.

We present the curve-to-curve (C2C) distance~\cite{zhang18} as the first evaluation metric. For each point on the predicted path, the distance to the nearest point on the GT path is computed. They are averaged and it is defined as the distance from the predicted to the GT. The distance from the GT to the predicted is computed similarly, and the C2C distance is the average of those two distances. While it could indicate the coverage of the predicted path on the entire course of the small bowel, it is computed in disregard of the order of the tracking, and the error of crossing the walls. To better understand the tracking capability under the contact issue, we provide the maximum length of the GT path that is tracked without making an error. This metric can be of interest for image guided interventions that require figuring out the exact structure of the bowel.

\section{Results}


The proposed method, `TSP + Cyl', is compared with the baseline methods in Table~\ref{tab:quan_res}. It has been greatly benefited by the must-pass nodes in increasing the coverage of the tracking, which is implied by the large decrease in the C2C distance. It is further improved by the inclusion of the local cylinders fitting in terms of the another metric, the maximum length of the GT path that is tracked without making an error. While the new cost function that is facilitated by the cylinders could prevent the tracking from crossing walls at the contact area, it could occasionally let it keep the distance to the GT path rather than being zigzagging around the GT. In practice, the improvement in the maximum length measure has come with little increase in the C2C distance. We note that our GT path is not exact centerlines of the small bowel. Thus, a little more deviation from the GT path could not mean that much. As mentioned in Section~\ref{ssec:eval_details}, we believe the maximum length measure is more relevant in tracking the small bowel path.

Since the proposed method includes small bowel segmentation in it, its quality affects the path tracking outcome. By ruling out the negative effect of the segmentation error, the gain of the proposed method became more evident.

\begin{table}[t]
    \caption{Quantitative results for the baseline and the proposed methods. The curve-to-curve (C2C) distance and the maximum length of the GT path that is tracked without making an error are presented. 
    `SP' denotes computing the shortest path without the use of the must-pass nodes. `TSP' denotes finding the optimal path by solving the travelling salesman problem, where the must-pass nodes are considered. `TSP + Cyl' is the proposed method. The results of using GT segmentation in place of the predicted segmentation, are also presented.
    } 
    \label{tab:quan_res}
    \vspace{-0.5cm}
    \begin{center}
    \small
    \begin{tabular}{c|c|c|c}
    Method & Segm. & C2C (mm) & Max. len. w/o error (mm) \\
    \hline
    SP & Pred. & 28.0 $\pm$ 7.5 & 756.2 $\pm$ 341.5\\
    \hline
    TSP~\cite{shin21_sb_track} & Pred. & 4.2 $\pm$ 1.2 & 810.0 $\pm$ 193.6\\
    \hline
    TSP + Cyl & Pred. & 4.7 $\pm$ 1.3 & \textbf{863.8} $\pm$ 202.3\\
    \hline
    TSP~\cite{shin21_sb_track} & GT & 3.3 $\pm$ 0.3 & 840.6 $\pm$ 273.1\\
    \hline
    TSP + Cyl & GT & 3.8 $\pm$ 0.4 & \textbf{983.8} $\pm$ 285.8\\
    \end{tabular}
    \end{center}
    \vspace{-0.5cm}
\end{table}


Although our graph is constructed based on the small bowel wall detection, it is not enough to prevent the penetration over the wall. Indistinct appearance of the wall, exemplified in Fig.~\ref{fig:cyl} (A), causes false negatives in the detection. It finally could allow the tracking to cross over it. In Fig.~\ref{fig:cyl} (C), the cylinder was successfully estimated from the incomplete wall detection, and it is used as an advanced measure for preventing the penetration. Fig.~\ref{fig:qual_res} shows example path tracking results. The path that penetrates the walls is corrected in the result of the proposed method.

\section{Conclusion}

A new graph-based method for small bowel path tracking was presented. Two main things that make the path tracking difficult for the small bowel are the contact and the indistinct appearance of the wall. To make the tracking more robust in existence of those issues, we made use of the fact that the bowel is a tube. We fitted piecewise cylinders along its course, and guided the tracking to follow their orientation. The better tracked path provides a better understanding on its structure. Especially, it could help precise localization of diseases in the small bowel, and image-guided interventions where a device approaches according to the identified structure. In future work, we plan to incorporate a measure for making the framework less affected by the segmentation quality.

\section{Compliance with ethical standards}

IRB approval has been obtained by the local Ethics Committee for the dataset used.



\section{Acknowledgments}
 
We thank Dr. James Gulley for patient referral and for providing access to CT scans. This research was supported by the Intramural Research Program of the National Institutes of Health, Clinical Center.


\end{document}